\documentclass[runningheads, anonymous, UTF8]{llncs}
\usepackage[T1]{fontenc}

\usepackage{subfigure} 

\usepackage[table]{xcolor}
\usepackage{colortbl}
\usepackage{xeCJK} 
\usepackage{graphicx}
\usepackage{xcolor}
\usepackage{graphicx}
\usepackage{subcaption}
\usepackage{amsmath}
\usepackage{colortbl}
\usepackage{algorithm}
\usepackage{algorithmic}
\usepackage{amssymb} 
\usepackage{marvosym}
\usepackage{booktabs}
\usepackage{threeparttable}
\usepackage{tabularx}

\usepackage{setspace}  
\usepackage{enumitem}  
\usepackage{arydshln} 

\usepackage{url}

\usepackage[
  style=numeric,
  maxnames=3, 
  minnames=1, 
  backend=biber,
  sorting=none,         
  citestyle=numeric-comp
]{biblatex}

\DeclareFieldFormat{labelnumberwidth}{#1.}
\DeclareFieldFormat{shorthandwidth}{#1.}

\DeclareFieldFormat{citesuperscript}{%
  \mkbibsuperscript{#1}}
\DeclareFieldFormat{cite}{#1}
\DeclareFieldFormat{parencite}{#1}

\DeclareFieldFormat[article]{volume}{\textbf{#1}}

\definecolor{gold}{RGB}{255,230,0}
\definecolor{silver}{RGB}{192,192,192}
\definecolor{bronze}{RGB}{205,127,50}

\begin{document}
\title{VisualQuest: A Benchmark for Abstract Visual Reasoning in MLLMs}

\titlerunning{VisualQuest: A Benchmark for Abstract Visual Reasoning in MLLMs}

\author{Kelaiti Xiao\inst{1,3} \and
Liang Yang\inst{1}\and
PAERHATI Tulajiang\inst{1,3} \and
Dongyu Zhang \inst{2} \and
Hongfei Lin \inst{1} \textsuperscript{(\Letter)}
}

\authorrunning{K. Xiao et al.}
\institute{School of Computer Science and Technology, Dalian University of Technology, Dalian, China \\
\{xklt@mail.,liang@,prht@mail.,zhangdongyu,hflin@\}dlut.edu.cn
\and
School of Foreign Languages, Dalian University of Technology, Dalian, China
\and
School of Computer Science and Technology, Xinjiang Normal University, Urumqi, China
}

\maketitle              
\begin{abstract}
We introduce VisualQuest, a novel dataset designed to rigorously evaluate multimodal large language models (MLLMs) on abstract visual reasoning tasks that require integration of symbolic, cultural, and linguistic knowledge. Unlike existing benchmarks focused on direct image captioning or classification of realistic images, VisualQuest comprises 3,551 non-photographic, stylized images spanning four categories—Public Figures, Popular Culture, Linguistic Expressions, and Literary Works—each paired with targeted questions to probe complex reasoning. We benchmark ten state-of-the-art MLLMs, revealing that only Gemini-2.5-flash and GPT-4o achieve strong overall performance, while 3.7\% of images remain unrecognized by any model, underscoring persistent challenges in multimodal understanding. Fine-grained analysis shows Gemini excels in recognizing stylized public figures, whereas GPT-4o leads in linguistic reasoning tasks such as visual puns and emoji combinations. VisualQuest thus provides a comprehensive and challenging resource for advancing research in abstract visual reasoning and highlights key areas for future model improvement. The Dataset is available at https://github.com/xkt88/VISUALQUEST
\end{abstract}

\section{Introduction}
Large Language Models (LLMs) have demonstrated remarkable success in processing complex textual data, and their capabilities have recently extended into multimodal tasks that integrate visual and textual inputs\cite{wu2023multimodal}. Applications such as automated captioning of everyday scenes and the analysis of medical imagery have showcased these models’ enhanced capacity to interpret visual information when combined with robust learning paradigms\cite{Yin2023A}. Nevertheless, most existing multimodal benchmarks predominantly focus on direct image captioning or coarse-grained classification tasks using real-world datasets like ImageNet\cite{imagenet}. While valuable, these approaches do not adequately assess the ability of MLLMs to leverage domain-specific background knowledge in more sophisticated recognition scenarios.

\begin{figure}
    \centering
    \includegraphics[width=\linewidth]{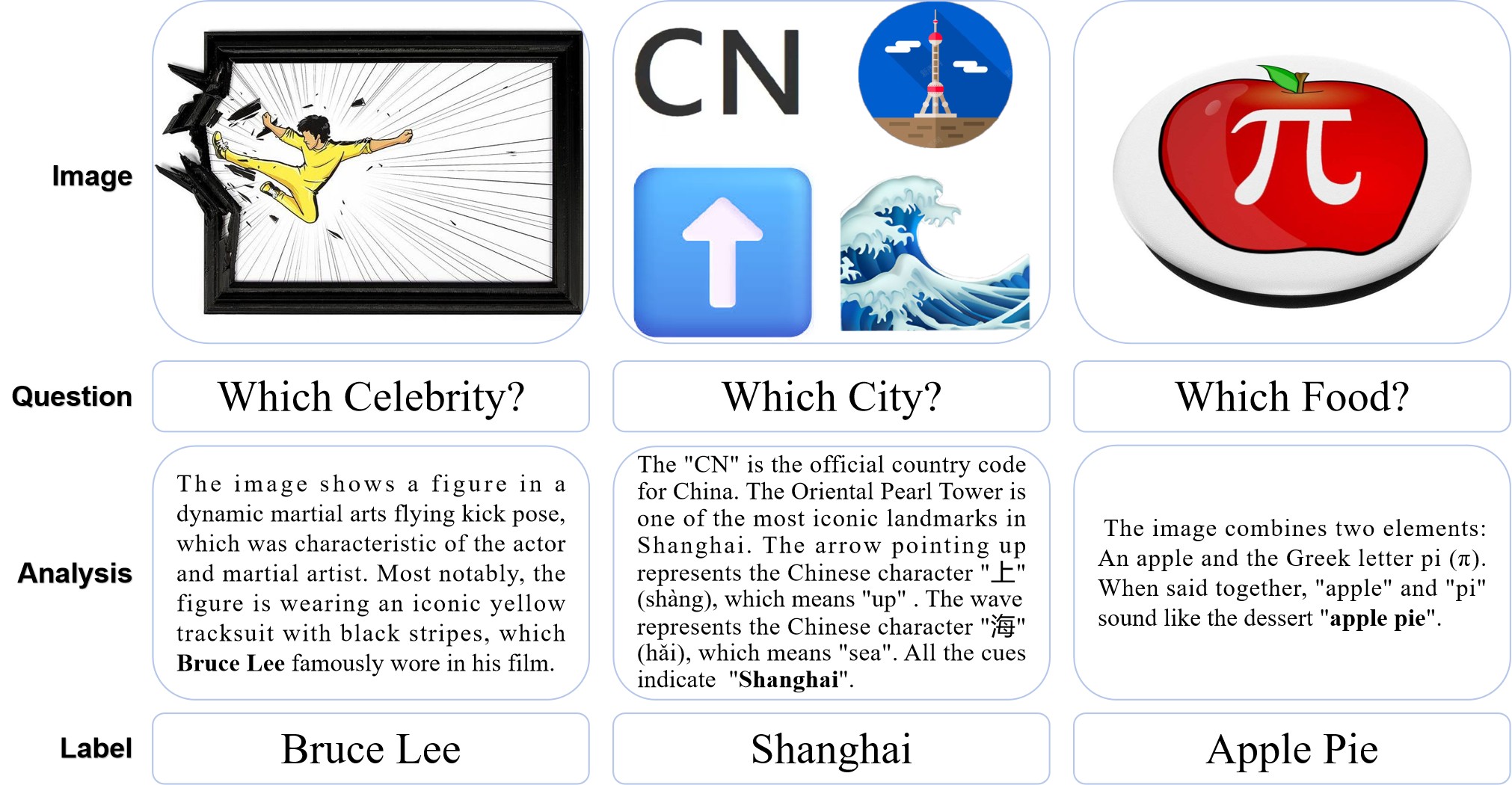}
    \caption{Examples from the VisualQuest dataset illustrating the task format. For each input image and question (top two rows), the model must generate a reasoning analysis and the correct label (bottom two rows)}
    \label{img1}
\end{figure}

To illustrate the unique challenges posed by abstractive images, consider the examples shown in Figure 1. Rather than presenting straightforward photographs, these images require sophisticated interpretation through multiple layers of abstraction. The first example depicts a figure in a dynamic martial arts pose wearing an iconic yellow tracksuit—visual cues that, combined with cultural knowledge, point to Bruce Lee. The second example employs symbolic representation: the country code "CN", Shanghai's Oriental Pearl Tower, and Chinese characters "上" (shàng, meaning "up") and "海" (hǎi, meaning "sea") that phonetically spell "Shanghai." The third example demonstrates a visual pun where an apple paired with the Greek letter \(\pi\) creates the phonetic representation of "apple pie."  These examples highlight how abstractive images challenges models to move beyond simple object detection and engage in complex reasoning that integrates visual symbolism, cultural references, linguistic knowledge, and abstract thinking. Such capabilities are essential for human-like visual understanding, yet few datasets have been specifically designed to evaluate VLMs on these multifaceted capabilities.

To bridge this gap, we introduce “VisualQuest,” a dataset specifically designed to evaluate the performance of VLMs on images that deviate from conventional photographic representations. Unlike traditional image datasets, VisualQuest centers on visuals infused with stylized, symbolic, and at times surrealistic elements.

The main contributions of this paper are as follows:

\begin{itemize}
    \item We introduce VisualQuest, a novel dataset of 3,551 abstractive images spanning four categories with various non-realistic presentation methods, specifically designed to evaluate MLLMs' abstract visual reasoning capabilities beyond conventional photographic representations .
    
    \item We conduct a comprehensive evaluation of 10 state-of-the-art MLLMs, revealing that only Gemini-2.5-flash and GPT-4o achieve strong performance, while 3.7\% of images remain unrecognized by all models, highlighting persistent challenges in multimodal understanding .
    
    \item We provide fine-grained analysis across background knowledge categories and presentation methods, uncovering distinct model strengths: Gemini-2.5-flash demonstrates exceptional visual recognition capabilities particularly for stylized public figures, while GPT-4o excels in linguistic reasoning tasks including visual puns and emoji combinations.
\end{itemize}

\section{Related Works} \label{sec:format}
Table 1 provides an overview of major visual reasoning datasets developed in recent years, highlighting the breadth and evolution of benchmarks in this field. Many of these datasets build upon one another, reflecting a strong tradition of legacy and cumulative progress in visual reasoning research. This section reviews the key contributions and interconnections among these datasets, setting the stage for understanding current challenges and motivating the need for further advancements in multimodal visual reasoning.

\begin{table}[ht]
\small
\setlength{\tabcolsep}{4pt}
\centering
\caption{Existing Visual Reasoning Datasets}
\label{tab:llm_counts}
\begin{tabular}{@{}lllc@{}}
\toprule
\textbf{Dataset} & \textbf{Size} & \textbf{Origin} & \textbf{Source/Year} \\
\midrule
SHAPES\cite{SHAPES} &  15,616 & Colored Shapes & CVPR 2016 \\
Visual Genome\cite{genome} & 108k &  YFCC100M\cite{YFCC} \& MS COCO\cite{MSCOCO} & IJCV 2017\\
FVQA\cite{FVQA} & 2,190& MS COCO\cite{MSCOCO} & PAMI 2017 \\
CLEVR\cite{CLEVR}& 100k & Geometrical objects & CVPR 2017\\
VCR\cite{VCR} & 110k & Movie Scenes & CVPR 2019 \\
GQA\cite{GQA} & 113k& Common Objects & CVPR 2019 \\
RAVEN\cite{RAVEN} & 1,120k & Raven's Matrices & CVPR 2019\\
NLVR\cite{NLVR} &  92,244 & Geometrical Shapes & ACL  2019 \\
KRVQA\cite{KRVQA} & 32,910& Visual Genome\cite{genome} \& FVQA\cite{FVQA} & TNNLS 2022 \\
CVR\cite{CVR} & 12,500 & Generated Geometric Shapes & NIPS 2022 \\
QLEVR\cite{qlevr} & 100k & Generated geometrical objects & NAACL 2022 \\
ConceptARC\cite{conceptarc} & 480 & Geometrical Shapes & Nature 2023 \\
VSR\cite{VSR} & 10,972 & MS COCO\cite{MSCOCO} & TACL 2023 \\
Super-CLEVR\cite{super} & 30k & Vehicles from UDA-Part\cite{UDA} & CVPR 2023 \\
Oogiri-Go\cite{oogiri} & 130k & Oogiri game  & CVPR 2024\\
MCQA\cite{PRCV} & 22,860 & Charts from real-world data & PRCV 2024 \\
MATH-V\cite{mathvision} & 3,040 &  Real Math Competitions & NIPS 2024 \\
VISREAS\cite{visreas} & 113k &  Visual Genome\cite{genome} & ACL 2024 \\
VisuLogic\cite{visulogic} & 1,000 & Visual Logic Puzzles & Arxiv 2025 \\
VGRP-Bench\cite{VRGP} & 2,000k & Visual Grid Reasoning Puzzles & Arxiv 2025 \\
\bottomrule
\end{tabular}
\end{table}

\subsection{Compositional and Spatial Visual Reasoning}

A substantial body of work has focused on evaluating models' fundamental visual reasoning capabilities, particularly in understanding spatial relationships and compositional structures. Early efforts like the SHAPES\cite{SHAPES} dataset by Andreas et al. used synthetic images of colored shape arrangements to test compositional reasoning abilities. This synthetic approach was further refined in datasets like CLEVR\cite{CLEVR} by Johnson et al. which employed 3D-rendered scenes to evaluate spatial relations, counting, and logical operations, and NLVR\cite{NLVR} by Suhr et al. which grounded natural language statements in geometric arrangements to test set-theoretic and spatial reasoning. More recent work has shifted toward real-world complexity, with Hudson and Manning's GQA\cite{GQA} dataset leveraging Visual Genome scene graphs to evaluate spatial relations and compositional reasoning in natural images, while Liu et al.'s VSR\cite{VSR} dataset specifically targeted 66 types of spatial relations in MS COCO images. Recent synthetic datasets like CVR\cite{CVR} by Zerroug et al. and VISREAS\cite{visreas} by Akter et al. have emphasized compositional efficiency and multi-hop spatial reasoning, demonstrating continued interest in understanding how models leverage compositionality for visual reasoning tasks.

\subsection{Abstract, Pattern-Based, and Mathematical Reasoning}

Another line of research has explored abstract reasoning and mathematical understanding in visual contexts. Zhang et al.'s RAVEN\cite{RAVEN} dataset adapted Raven's Progressive Matrices to evaluate structural, relational, and analogical reasoning using gray-scale geometric patterns, while Mitchell et al.'s ConceptARC\cite{conceptarc} drew inspiration from Chollet's ARC benchmark to test abstract concept understanding through colored grid puzzles. Ren et al.'s VGRP-Bench\cite{VRGP} synthesized 20 diverse visual grid reasoning puzzles to comprehensively evaluate pattern matching and logical reasoning capabilities. In parallel, several datasets have targeted mathematical and quantitative reasoning: Li and Søgaard's QLEVR\cite{qlevr} focused on understanding 27 different quantifiers and set relationships in synthetic 3D scenes, Wu et al.'s MCQA\cite{PRCV} evaluated complex logical reasoning across multiple charts, and Wang et al.'s MATH-Vision\cite{mathvision} collected real-world mathematical competition problems to test multimodal mathematical reasoning across 16 disciplines including algebraic, geometric, and combinatorial reasoning.

\subsection{Knowledge-Grounded and High-Level Cognitive Reasoning}

Beyond direct visual analysis, several datasets have explored the integration of external knowledge with visual reasoning. Wang et al.'s FVQA\cite{FVQA} and Cao et al.'s KRVQA\cite{KRVQA} specifically require models to leverage commonsense knowledge and external knowledge bases for visual question answering. High-level cognitive reasoning has been addressed through datasets like VCR\cite{VCR} by Zellers et al. which uses movie scenes to evaluate understanding of human actions, mental states, and social dynamics, and Oogiri-GO\cite{oogiri} by Zhong et al. which uniquely tests creative thinking and humor generation across multiple languages. For comprehensive visual understanding, Krishna et al.'s Visual Genome\cite{genome} provides extensive annotations for real-world visual reasoning across multiple dimensions, while Xu et al.'s VisuLogic\cite{visulogic} covers six categories of vision-centric logical reasoning tasks. Finally, Li et al.'s Super-CLEVR\cite{super} specifically addresses the critical issue of domain robustness, evaluating how models generalize across variations in visual complexity, question redundancy, concept distribution, and compositionality.

Despite the impressive progress and diversity of existing visual reasoning datasets, a key gap remains: most benchmarks are narrowly focused on specific data types, with limited and highly concentrated background knowledge requirements. Furthermore, prior work has largely relied on uniform or realistic presentation methods, without systematically exploring how varied and stylized visual representations impact reasoning. To address these limitations, our work introduces a new dataset specifically designed to evaluate models’ abstractive visual reasoning abilities using images that require extensive background knowledge and span multiple non-realistic presentation methods, thus providing a more comprehensive and challenging benchmark for multimodal understanding.

\section{Dataset Details}

\begin{figure}
    \centering
    \includegraphics[width=\linewidth]{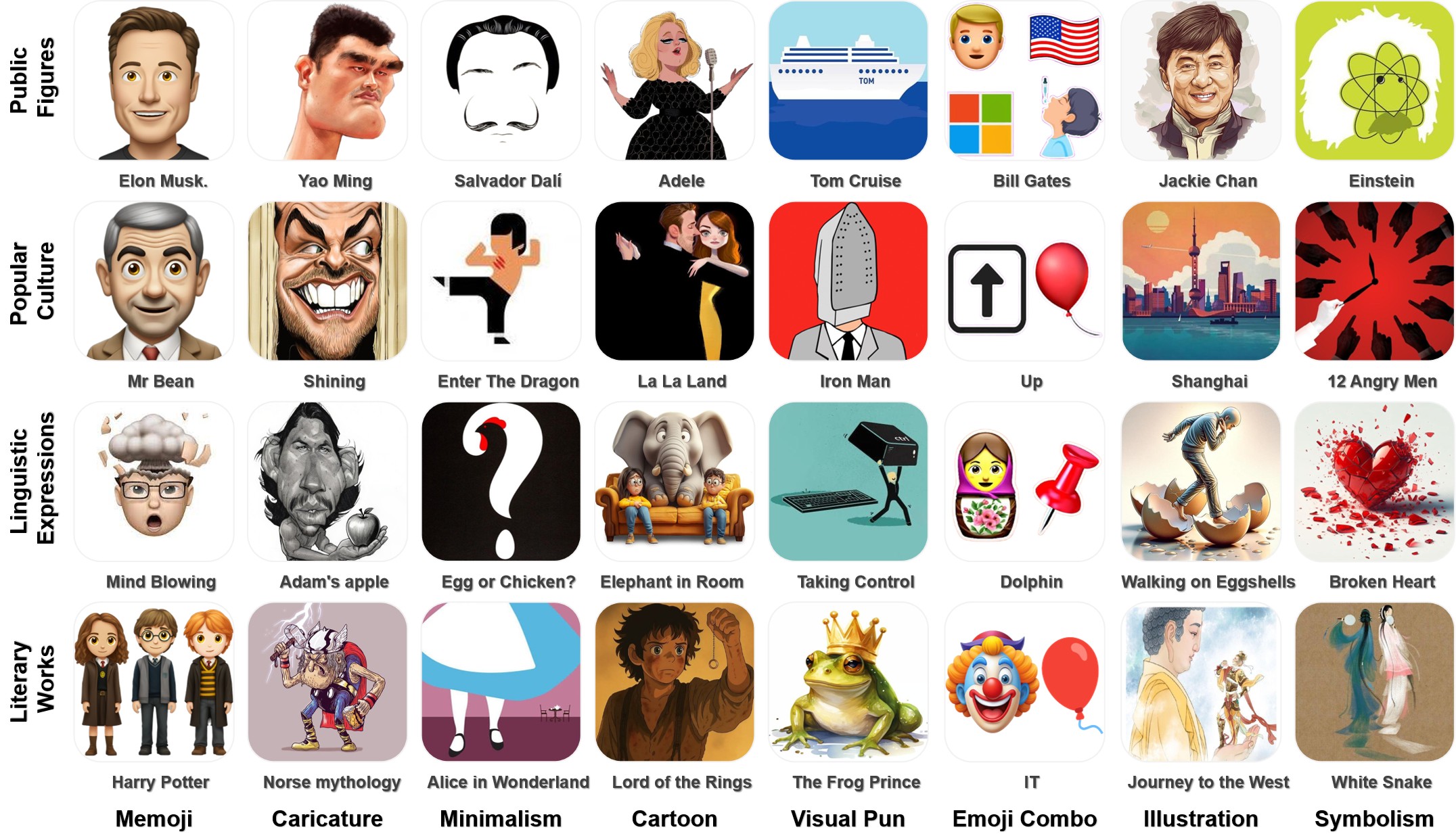}
    \caption{Example Set for Training}
    \label{img1}
\end{figure}

\label{sec:dataset_details}
The following section provides a comprehensive overview of the VisualQuest dataset, detailing its construction, organization, and unique features. We describe the systematic process used for data collection and curation, outline the dataset’s categorical structure and presentation methods, and explain the label schema adopted for each image. This foundation is essential for understanding the subsequent experimental evaluations and the specific challenges posed by VisualQuest’s emphasis on abstract, symbolic, and non-realistic visual reasoning.

\subsection{Data Collection} \label{subsec:data_collection}
To ensure high-quality and consistent data collection, we implemented a structured, multi-stage process for recruiting and training student collectors:
\begin{itemize} \item
\textbf{Training:} We provided prospective collectors with a curated set of example images (Fig. 2) that illustrated the range of stylized, symbolic, and non-traditional depictions we sought. These examples spanned all four target categories and various presentation methods, allowing collectors to internalize the desired concepts and visual cues. 
\item \textbf{Testing and Selection:} After the training phase, we evaluated each collector using 32 additional test images to verify their ability to accurately identify and classify images according to our criteria. Based on their performance scores, we selected the top 10 collectors to proceed with the data collection task.
\item \textbf{Main Collection:} Each of the 10 selected collectors was then instructed to gather 500 images from diverse online resources, adhering to the following criteria: \begin{itemize}
\item Images must belong to one of four targeted categories: \emph{Public Figures, Popular Media, Linguistic Expressions}, or \emph{Literature Works}. 
\item Images must use \textbf{non-realistic} or stylized presentation methods—such as \emph{caricature}, \emph{illustration}, \emph{emoji combination}, or \emph{visual pun}. 
\item Images must include sufficient cultural, literary, or symbolic elements that deviate from straightforward photographic representations. \end{itemize} \end{itemize}

This systematic approach—from training through testing to final collection—helped ensure consistency and quality in the collected data, aligning all contributors with the project's emphasis on abstract and symbolic visual reasoning.

\subsection{Data Curation and Annotation}
\label{subsec:filtering}

Starting from an initial pool of 5{,}000 images, we applied a series of filtering and annotation steps to ensure dataset quality and consistency, resulting in a final set of 3{,}551 images:

\begin{itemize}
    \item \textbf{Duplicate and Unrecognizable Image Removal:} All repeated images were eliminated to ensure uniqueness. Next, ten English-major reviewers independently attempted to identify the main entity or concept in each image. Images that failed to achieve agreement from at least two reviewers were excluded, ensuring that only interpretable visuals remained.
    
    \item \textbf{Presentation Method Annotation:} Each image was labeled with a presentation style, initially selected from eight canonical categories, with an ``Other'' option for novel styles. After review, new categories such as ``Iconography'' and ``Surrealism'' were added for a small number of images. The final label for each image was determined by majority vote, with ties resolved by a designated arbiter.
    
    \item \textbf{Removal of Non-Essential Text:} Images containing superimposed text (e.g., names or titles) that could bias recognition were edited to remove non-essential characters. However, text integral to the image’s symbolism or pun was retained.
    
    \item \textbf{Question Annotation:} Each image was paired with a concise query, beginning with ``Which'' for straightforward recognition or ``Guess'' for puzzle-like images requiring deeper inference. This structure encourages LLMs to combine recognition, domain knowledge, and reasoning.
    
    \item \textbf{Label Standardization:} Label usage, presentation method names, and case conventions were harmonized across the dataset. For images with multiple valid answers (e.g., character name or literary source), any correct answer was accepted. This step ensured consistency throughout the dataset.
\end{itemize}

\subsection{Dataset Composition}

\begin{table}
\caption{\small VisualQuest dataset composition by category and presentation method}\label{tab:dataset_composition}
\begin{center}
\begin{tabular}{|l|p{8.5cm}|}
\hline
\textbf{Category} & \textbf{Description} \\
\hline
Public Figures & 1,024 images including authors, actors, scientists, and other notable personalities \\
\hline
Popular Culture & 690 images featuring films, television shows, anime, video games, and iconic cities \\
\hline
Linguistic Expressions & 954 images representing idioms, proverbs, and phrases \\
\hline
Literary Works & 883 images depicting novels, fairy tales, fictional characters, and deities \\
\hline
Presentation Method & Memoji (89), Caricature (959), Minimalism (149), Cartoon (45), Iconography (1), Surrealism (4), Visual Pun (692), Emoji Combination (655), Illustration (829), and Symbolism (128) \\
\hline
\end{tabular}
\end{center}
\end{table}
\label{subsec:dataset_structure}

As shown in Table~\ref{tab:dataset_composition}, the VisualQuest dataset is organized into four major categories, each emphasizing a distinct type of background knowledge:
\begin{itemize}
    \item \textbf{Public Figures:} Focuses on classical and widely recognized personalities, requiring models to draw on long-standing cultural and historical knowledge.
    \item \textbf{Popular Culture:} Emphasizes recent developments in media such as films, television, anime, video games, and iconic cities, testing models' ability to accommodate contemporary trends and references.
    \item \textbf{Linguistic Expressions:} Contains images representing idioms, proverbs, and phrases, challenging models to interpret figurative language and abstract verbal concepts through visual cues.
    \item \textbf{Literature Works:} Includes depictions of novels, fairy tales, fictional characters, and deities, assessing familiarity with both canonical literature and imaginative domains.
\end{itemize}

Importantly, all images in VisualQuest use non-realistic presentation methods, focusing on stylized and abstract visual forms. This design moves beyond simple object recognition, requiring models to integrate symbolic, cultural, and artistic reasoning. While the dataset is not strictly balanced across categories or presentation methods, its structure provides a strong foundation for future extensions and encourages the community to further expand and enrich the dataset for multimodal reasoning research.

\subsection{Label structure}

\begin{table}
\caption{\small Example of Image Label Structure}\label{tab:image_data}
\begin{center}
\begin{tabular}{|l|l|}
\hline
\textbf{\;Attribute} & \textbf{\;Sample Value} \\
\hline
\;Image ID & \;"0118": \\
\hline
\;Ground Truth & ["Bruce Lee; Li Xiaolong", \\
\hline
\;Category & \;"Public Figures", \\
\hline
\;Question\; & \;"Which Celebrity?", \\
\hline
\;Presentation Method\; & \;"Caricature", \\
\hline
\;Recognition Vector & \;[0, 1, 1, 1, 1, 0, 1, 0, 0, 0],\; \\
\hline
\;Recognition Count & \;5 ]\\
\hline
\end{tabular}
\end{center}
\end{table}

As shown in Table~\ref{tab:image_data}, each image in the VisualQuest dataset is stored in a dictionary-like structure, where the "Image ID" serves as the key and the associated list of attributes forms the value. The label structure is organized as follows:

\begin{itemize}
    \item \textbf{Image ID}: Each image is uniquely identified by a string key ranging from "0000" to "3551", corresponding to the 3,551 images in the current dataset.
    \item \textbf{Ground Truth}: This field contains one or more valid labels for the image. For example, image "0118" accepts both "Bruce Lee" and "Li Xiaolong" as correct answers, reflecting the possibility of multiple ground truths per image.
    \item \textbf{Category}: The image's semantic category, such as "Public Figures", is specified here. The full list of categories is detailed in Section~3.3, and the taxonomy can be expanded or refined in future dataset extensions.
    \item \textbf{Question}: Each image is paired with a specific question, which not only guides the model on the recognition task but also helps indicate the image's subcategory or intended reasoning challenge.
    \item \textbf{Recognition Vector}: This binary vector records the recognition results from 10 baseline MLLMs, with the order of models defined in Table~4. The first entry corresponds to \textit{GPT-4o}, and the last to \textit{Phi-4-multimodal-instruct}.
    \item \textbf{Recognition Count}: This integer indicates how many of the 10 MLLMs successfully recognized the image, with possible values ranging from 0 (none recognized) to 10 (all recognized).
\end{itemize}

\section{Experiments}
This section details the selection of models, the benchmarking methodology, and the comparative results across various dimensions, including overall recognition accuracy, performance by background knowledge category, and effectiveness across different visual presentation methods.
\subsection{Evaluated Multimodal Large Language Models}
\begin{table}
\caption{\small Overview of MLLMs}\label{tab:llm_overview}
\begin{center}
\begin{tabular}{|l|c|c|c|}
\hline
\textbf{LLM} & \textbf{Availability} & \textbf{Company} & \textbf{Released Date} \\
\hline
GPT-4o & Closed-source & OpenAI & March 26, 2025 \\
Claude-Sonnet-4 & Closed-source & Anthropic & May 22, 2025 \\
Gemini-2.5-flash  & Closed-source & Google & April 17, 2025 \\
Grok-3 & Closed-source & xAI & February 17, 2025 \\
Doubao-1.5-vision-pro & Closed-source & ByteDance & January 22, 2025\\
Gemma-3-27b-it & Open-source & Google & March 12, 2025 \\
GLM-4V-Plus  & Closed-source & THUDM & August 28, 2024 \\
Llama-4-Maverick & Open-source & Meta & April 5, 2025 \\
QVQ-72B-Preview & Open-source & Alibaba & December 25, 2024 \\
Phi-4-multimodal-instruct   & Open-source & Microsoft & February 26, 2025 \\
\hline
\end{tabular}
\end{center}
\end{table}

As shown in Table~\ref{tab:llm_overview}, our evaluation covers 10 prominent multimodal large language models (MLLMs), including six closed-source and four open-source models. For each company, we selected the most recent and widely adopted model version, deliberately excluding specialized "chain of thought" reasoning variants (such as GPT-o1, DeepSeek-R1) to ensure a fair comparison. This approach allows us to focus on mainstream MLLMs that strike a balance between cost, response latency, and overall performance.

Regarding model access, we leveraged the Deepinfra\footnote{https://deepinfra.com/} platform for Claude, Gemini, Gemma, Llama, QVQ, and Phi, due to their availability and convenience on this service. The remaining MLLMs, which were not accessible via Deepinfra, were evaluated using their official APIs. This setup ensures consistency and practicality in benchmarking across a diverse set of leading MLLMs.

\subsection{Overall Model Performance}

\begin{figure}[htbp]
    \centering
    \subfigure[]{
        \includegraphics[width=0.38\linewidth]{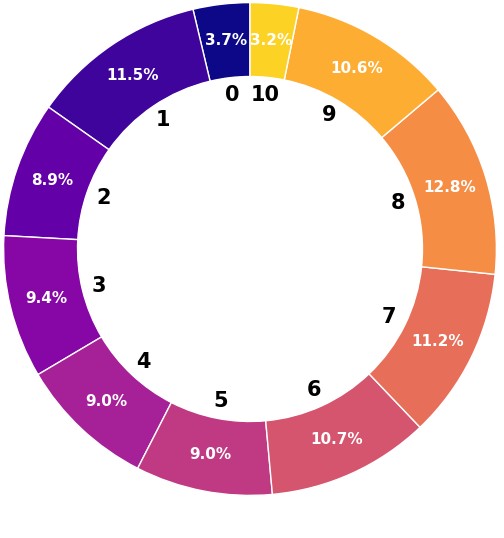}
        \label{fig:donut}
    }
    \subfigure[]{
        \includegraphics[width=0.57\linewidth]{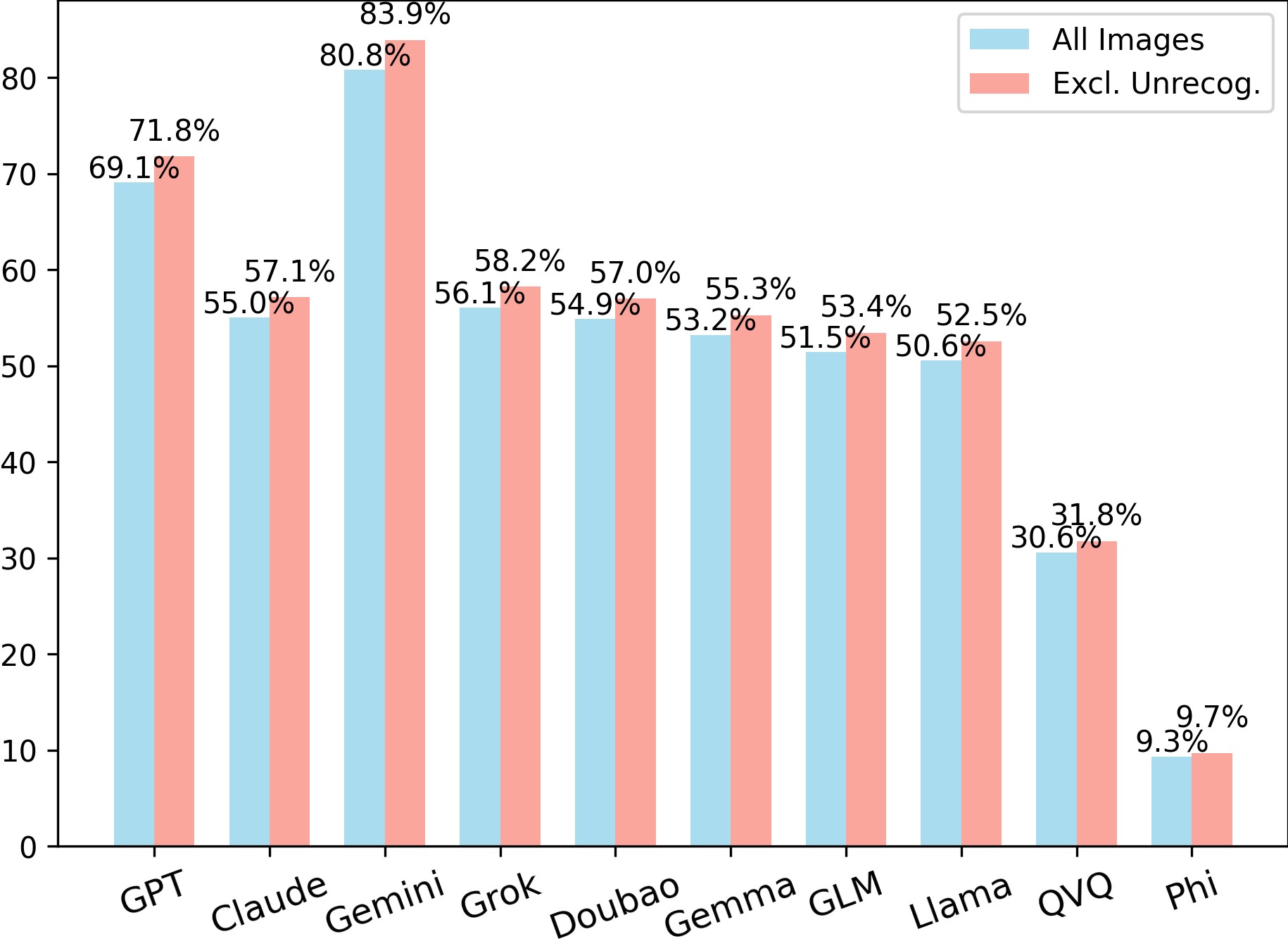}
        \label{fig:bar}
    }
    \caption{ \small (a) Donut chart showing the distribution of images in the VisualQuest dataset by the number of MLLMs that recognized them. (b) Bar chart comparing the recognition accuracy of ten MLLMs on all images (blue) and after excluding universally unrecognized images (red).}
    \label{fig:side-by-side}
\end{figure}

Fig.~\ref{fig:side-by-side} provides a comprehensive overview of the recognition performance of ten leading MLLMs on the VisualQuest dataset. In Fig.~\ref{fig:donut}, the donut chart illustrates the distribution of images according to the number of MLLMs that successfully recognized them. Notably, 3.7\% of the images were not recognized by any of the ten models, while only 3.2\% were correctly identified by all models. The subset of images that none of the MLLMs could recognize is particularly significant, as these cases often represent instances where humans can reach consensus but current models universally fail. Such images offer valuable insights into the limitations of existing MLLMs and serve as promising seed data for future research on "easy image tests that MLLMs fail"\cite{conceptarc}.

Turning to Fig.~\ref{fig:bar}, we compare the recognition accuracy of each MLLM across all images (blue bars) and after excluding the universally unrecognized images (red bars). Gemini stands out as the top performer, being the only model to surpass the 80\% accuracy threshold. GPT, while trailing Gemini by approximately 10\%, remains the clear second-best model overall. Claude, Grok, Doubao, Gemma, GLM, and Llama form a middle tier, with accuracies ranging from 50\% to 55\%. QVQ lags behind this group, and Phi exhibits the lowest performance, with an overall accuracy below 10\%. These results underscore that, even on abstractive visual reasoning tasks, only Gemini and GPT demonstrate strong recognition capabilities. 

However, the existence of images that remain unrecognized by all current MLLMs indicates persistent challenges in multimodal understanding. Expanding and diversifying the image dataset will be crucial for driving further advances in this field and for developing models that can approach human-level consensus on complex visual reasoning tasks.

\subsection{Performance by Background Knowledge Category}

\begin{figure}
    \centering
    \includegraphics[width=\linewidth]{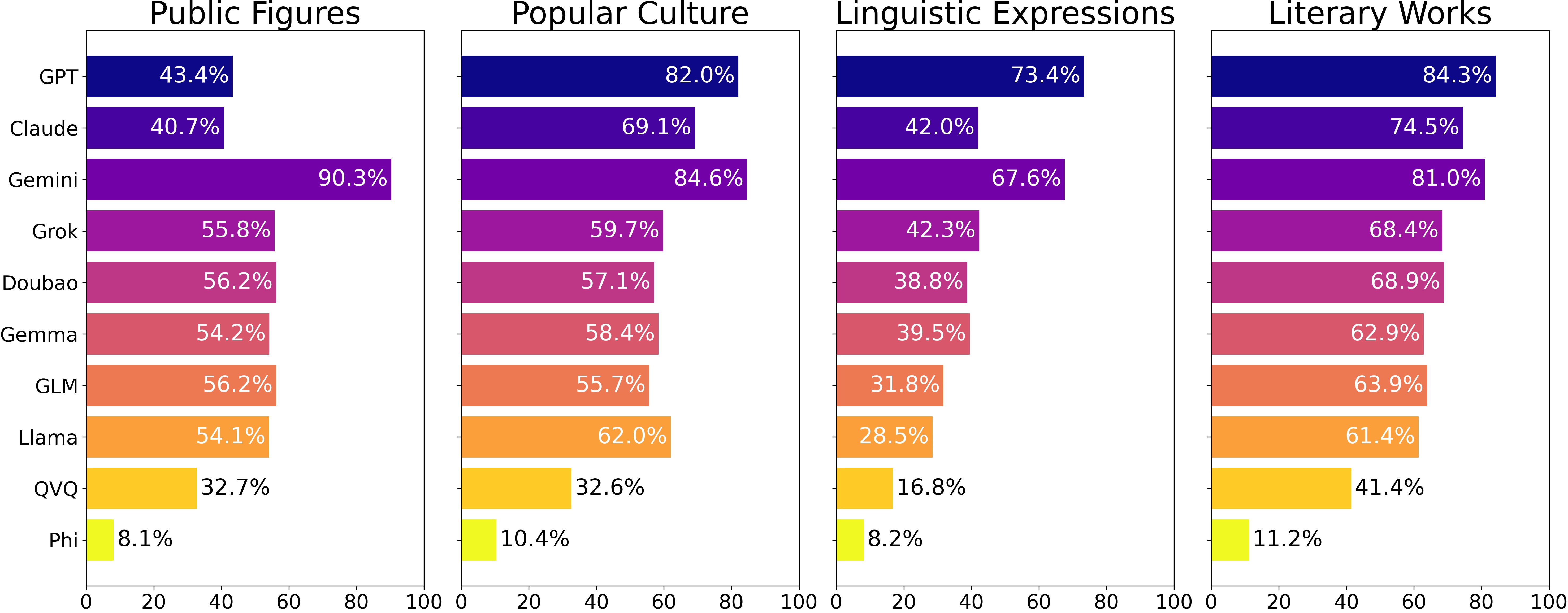}
    \caption{Performance of MLLMs on each Category}
    \label{img4}
\end{figure}

Figure~\ref{img4} presents the recognition accuracy of various MLLMs across the four general categories defined in Section 3.3, each of which is primarily distinguished by the type of background knowledge required for successful image recognition. These categories—Public Figures, Popular Culture, Linguistic Expressions, and Literary Works—were designed to probe different facets of factual, cultural, and linguistic reasoning. The results in Fig.~\ref{img4} thus provide a detailed breakdown of how each model leverages background knowledge in distinct recognition scenarios.

A striking observation from the "Public Figures" category is the substantial performance gap between Gemini and GPT. While Gemini achieves an impressive 90.3\% accuracy, GPT lags significantly behind at only 43.4\%, less than half of Gemini’s score. This disparity is the largest among all categories and is the primary contributor to the overall 10\% performance margin between the two models observed in Section 4.2. In contrast, for the other three categories—Popular Culture, Linguistic Expressions, and Literary Works—GPT and Gemini perform at a comparable level, with both models exceeding 80\% accuracy in Literary Works and Popular Culture, and maintaining similar results in Linguistic Expressions. These findings highlight that Gemini’s superior ability to recognize stylized depictions of public figures is a key differentiator in overall performance.

Interestingly, in the "Linguistic Expressions" and "Literary Works" categories, which demand strong literal language reasoning and abstract verbal understanding, GPT actually outperforms Gemini, achieving 73.4\% and 84.3\% accuracy, respectively. This suggests that GPT’s advanced language modeling capabilities provide a tangible advantage in tasks that require nuanced linguistic inference. However, it is notable that all models except GPT exhibit their lowest performance in the "Linguistic Expressions" category, indicating that, even after successful visual recognition, linguistic reasoning remains a significant challenge for most MLLMs. This underscores the need for further research to enhance the integration of visual and linguistic reasoning in future multimodal models.

\subsection{Performance by Visual Presentation Method}

\begin{table}[ht]
\caption{\small MLLM performance across eight visual presentation methods. Gold cells indicate the best-performing model, silver cells the second-best, and bronze cells the third-best for each presentation method. Values represent accuracy percentages.}
\label{tab:vlm_illustration}
\centering
\begin{tabular}{|l|c|c|c|c|c|c|c|c|}
\hline
\textbf{MLLMs} & \textbf{Memo} & \textbf{Caric} & \textbf{Minim} & \textbf{Cart} & \textbf{Symb} & \textbf{Illus} & \textbf{VPun} & \textbf{EComb} \\
\hline
GPT     &  \cellcolor{white}50.56 & 44.42 & \cellcolor{silver}80.00 & \cellcolor{silver}72.48 & \cellcolor{silver}78.91 & \cellcolor{gold}84.92 & \cellcolor{gold}69.80 & \cellcolor{gold}83.66 \\
Claude  & 67.42 & 44.00 & 64.44 & 51.01 & 51.56 & \cellcolor{bronze}75.51 & 42.63 & \cellcolor{bronze}57.71 \\
Gemini  & \cellcolor{gold}92.13 & \cellcolor{gold}92.39 & \cellcolor{gold}88.89 & \cellcolor{gold}77.85 & \cellcolor{gold}82.81 & \cellcolor{silver}82.87 & \cellcolor{silver}65.03 & \cellcolor{silver}76.03 \\
Grok    & \cellcolor{bronze}78.65 & 55.47 & 60.00 & \cellcolor{bronze}61.07 & 54.69 & 69.84 & \cellcolor{bronze}43.21 & 49.01 \\
Doubao  & \cellcolor{silver}84.27 & 56.20 & 55.56 & 45.64 & 46.09 & 69.96 & 38.15 & 51.30 \\
Gemma   & 68.54 & 55.68 & 57.78 & 49.66 & 52.34 & 64.54 & 37.14 & 50.99 \\
GLM     & 43.82 & \cellcolor{silver}60.06 & \cellcolor{bronze}68.89 & 56.38 & 45.31 & 66.47 & 36.13 & 36.34 \\
Llama   & 73.03 & \cellcolor{bronze}56.62 & 64.44 & 55.70 & \cellcolor{bronze}61.72 & 63.81 & 31.21 & 38.47 \\
QVQ     & 43.82 & 32.12 & 46.67 & 32.21 & 28.91 & 42.82 & 19.08 & 22.14 \\
Phi     & 10.11 & 8.65  & 13.33 & 5.37  & 0.78  & 12.42 & 8.24  & 9.77  \\
\hline
\end{tabular}
\end{table}

Table~\ref{tab:vlm_illustration} reports the recognition accuracy of ten leading MLLMs across eight distinct non-realistic visual presentation methods. These methods—such as memoji, caricature, minimalism, cartoon, symbolism, illustration, visual pun, and emoji combination—are the most common types of stylized image expression in the VisualQuest dataset, as defined in Table~2 and exemplified in Figure~2. 

A general inspection of the results reveals that Gemini and GPT overwhelmingly dominate the top positions across all presentation methods. Gemini secures first place in five categories and second place in three, while GPT achieves first place in three and second place in another three. However, GPT's performance is notably weaker in the Memoji and Caricature categories—presentation styles that are primarily used for the "Public Figures" category—highlighting a specific challenge for GPT in recognizing stylized depictions of well-known personalities. 

Conversely, GPT exhibits clear strengths in the "Visual Pun" and "Emoji Combination" categories, outperforming all other models. This confirms previous findings that GPT's advanced language reasoning capabilities provide a tangible advantage when interpreting images requiring linguistic inference or the integration of symbolic cues. Meanwhile, models such as Claude, Grok, Doubao, GLM, and Llama each display notable strengths in certain presentation methods, demonstrating that while no single model is universally superior, different architectures may excel under specific stylized visual conditions.

\section{Conclusion}

In this paper, we introduced VisualQuest, a novel benchmark specifically designed to evaluate the abstract visual reasoning capabilities of MLLMs. By curating a dataset of 3,551 non-photographic, stylized images across four categories, we systematically assessed ten state-of-the-art MLLMs. Our experiments revealed that only Gemini-2.5-flash and GPT-4o achieved strong overall performance, with notable strengths in recognizing stylized public figures and interpreting linguistic visual puns, respectively. Importantly, 3.7\% of the images remained unrecognized by any model, highlighting persistent challenges in multimodal understanding and the need for further research in this area.

Despite these contributions, our study has several limitations. The current dataset, while diverse, could benefit from further expansion to cover a broader range of visual styles and cultural references. Additionally, we observed that some images were unrecognizable by all the models, but due to page constraints, we were unable to conduct a detailed analysis of the underlying causes for these failures. A more thorough investigation into why certain images are universally challenging would provide valuable insights for both dataset design and model development.

Looking ahead, our future work will focus on analyzing the currently unrecognizable images to better understand the factors contributing to their difficulty. By leveraging these insights, we aim to extend the VisualQuest dataset in a direction that prioritizes images which are easy for humans to recognize but remain challenging for MLLMs. This approach will help drive the development of more robust and human-aligned multimodal models, ultimately advancing the field of abstract visual reasoning.

\section*{Acknowledgments}
This work is partially supported by the National Natural Science Foundation of China (No.62376051).

\begin{credits}

\end{credits}

\printbibliography

\end{document}